\pdfoutput=1
\documentclass{article}

\PassOptionsToPackage{numbers,compress}{natbib}
\usepackage[preprint]{neurips_2026}

\usepackage[utf8]{inputenc}
\usepackage[T1]{fontenc}
\usepackage{hyperref}
\usepackage{url}
\usepackage{booktabs}
\usepackage{amsfonts}
\usepackage{amssymb}
\usepackage{amsmath}
\usepackage{nicefrac}
\usepackage{microtype}
\usepackage{xcolor}
\usepackage{graphicx}
\usepackage{pifont}
\usepackage{multirow}
\usepackage{wrapfig}

\newlength{\archcompareimageheight}

\newcommand{\oursNanoDefFull}{41.46}
\newcommand{\oursNanoDefRare}{41.64}
\newcommand{\oursNanoDefNonRare}{41.41}
\newcommand{\oursNanoKnownFull}{44.26}
\newcommand{\oursNanoKnownRare}{45.71}
\newcommand{\oursNanoKnownNonRare}{43.83}

\newcommand{\oursSmallDefFull}{43.19}
\newcommand{\oursSmallDefRare}{43.07}
\newcommand{\oursSmallDefNonRare}{43.23}
\newcommand{\oursSmallKnownFull}{46.17}
\newcommand{\oursSmallKnownRare}{47.02}
\newcommand{\oursSmallKnownNonRare}{45.92}

\newcommand{\oursXXLDefFull}{44.49}
\newcommand{\oursXXLDefRare}{43.21}
\newcommand{\oursXXLDefNonRare}{44.86}
\newcommand{\oursXXLKnownFull}{47.45}
\newcommand{\oursXXLKnownRare}{47.76}
\newcommand{\oursXXLKnownNonRare}{47.36}

\newcommand{\ablSixgroupFull}{40.69}
\newcommand{\ablSixgroupRare}{40.55}
\newcommand{\ablSixgroupNonRare}{40.73}
\newcommand{\ablOnegroupFull}{37.27}
\newcommand{\ablOnegroupRare}{39.04}
\newcommand{\ablOnegroupNonRare}{36.74}

\title{HOIBlender: Blending Lightweight Detection with Vision-Language Priors for Efficient Human-Object Interaction Detection}

\author{%
  Junwen Chen \quad Keiji Yanai \\
  Department of Informatics, The University of Electro-Communications, Tokyo, Japan \\
  \texttt{\{chen-j,yanai\}@mm.inf.uec.ac.jp}
}

\begin{document}

\maketitle

\begin{abstract}
Human-object interaction (HOI) detection requires grounding an interacting human-object pair and recognizing the verb that links them, often under severe long-tail supervision. Recent methods improve accuracy with stronger detectors and vision-language priors, but many still stack heavy transformer encoders, intricate denoising schedules, or post-hoc semantic calibration on top of the detector. We present \textbf{HOIBlender}, an efficient HOI detector named after its core design principle: blending detector-grounded visual tokens, spatial subject-object reasoning, and BLIP-2 semantic priors inside one lightweight decoding pipeline. HOIBlender builds on an RF-DETR/LW-DETR-style foundation with a DINOv2 backbone and selects top-$K$ image-conditioned tokens directly from the multi-scale projector as subject and object candidates, removing the dedicated encoder stage retained by prior HOI methods. A dual-stage decoder first stabilizes human-object geometry and then performs verb and HOI classification through progressive BLIP-2 prior fusion, with classifier weights initialized from BLIP-2 text embeddings for long-tail categories. Grouped-query training further enriches optimization without increasing inference cost. Across three model scales (Nano, Small, 2XL), HOIBlender consistently outperforms SOV-STG-VLA and Hybrid-SOV-VLA on HICO-DET, reaching $44.49$ Default Full mAP in only $9$ training epochs while maintaining competitive latency and parameter budgets. These results show that lightweight detection, structured spatial-semantic decoding, and deeply integrated vision-language priors can be blended into a single efficient HOI pipeline.
\end{abstract}

\section{Introduction}

Human-object interaction (HOI) detection extends scene understanding beyond category-level recognition by jointly localizing a human, an interacted object, and the verb that links them as a unified $\langle$human, object, verb$\rangle$ triplet~\cite{chao2018learning,gupta2015visual}. This task underpins human-centered applications such as assistive perception, surveillance, robotics, and instruction following, where a system must understand not only \textit{what} appears in a scene but also \textit{how} entities act on one another. Compared with conventional object detection, HOI detection requires joint reasoning over object semantics, pairwise geometry, fine-grained action evidence, and contextual cues, all under a long-tailed label distribution in which many interactions appear only a few times.

HOI detection has undergone a clear methodological shift. Early CNN-based two-stage detectors~\cite{gkioxari2018detecting,gao2018ican,qi2018learning,gupta2019no} relied on hand-crafted pair construction and multi-stream fusion of human, object, and spatial cues. With the rise of transformers~\cite{waswani2017attention,carion2020end,zhu2020deformable}, query-based one-stage detectors such as QPIC~\cite{tamura2021qpic}, HOTR~\cite{kim2021hotr}, HOI-Trans~\cite{zou2021end}, and CDN~\cite{chen2021reformulating} reformulated HOI detection as set prediction and simplified end-to-end learning. Later methods strengthened spatial modeling~\cite{Kim_2022_CVPR,chen2023qahoi,zhang2023pvic}, decomposed query semantics~\cite{chen2022parallel,Zhang_2022_CVPR,zhou2022human}, and injected vision-language knowledge from CLIP~\cite{radford2021learning} or BLIP-2~\cite{li2023blip,li2022blip} into interaction classification~\cite{liao2022gen,ning2023hoiclip,mao2024clip4hoi,yuan2022rlip,yuan2023rlipv2,cao2024detecting}. These advances pushed accuracy upward, but they also expose two persistent gaps. First, many HOI heads still inherit relatively heavy DETR-style encoders even as object detection has moved toward faster and more scalable designs such as RT-DETR~\cite{zhao2024detrs}, RF-DETR~\cite{robinson2025rf}, and LW-DETR~\cite{chen2024lw}. Second, vision-language priors are often introduced as auxiliary calibration heads or zero-shot branches, leaving the visual decoding path that produces the final interaction prediction only weakly coupled to semantic knowledge.

\begin{figure}[t]
  \centering
  \settoheight{\archcompareimageheight}{\includegraphics[width=0.32\linewidth]{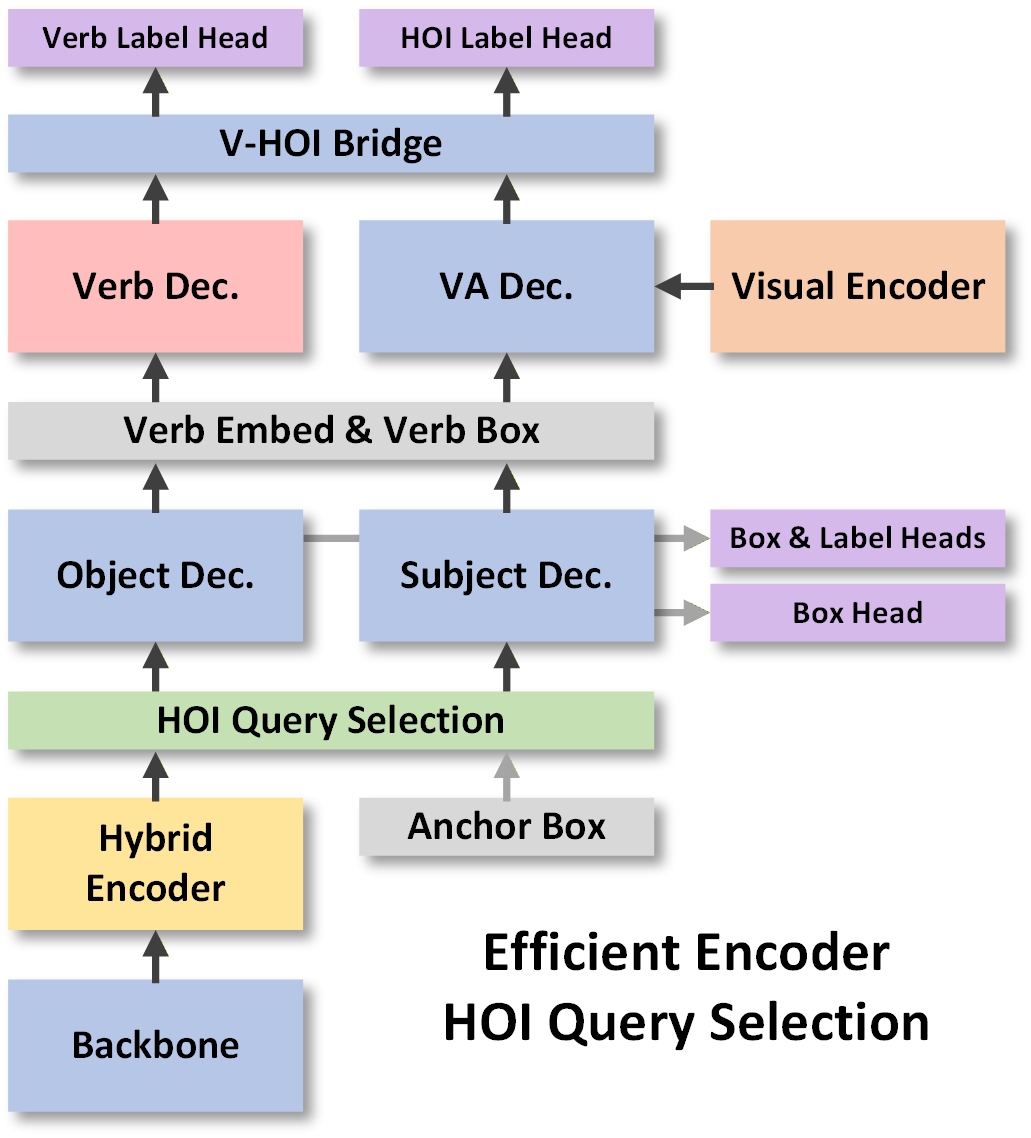}}
  \begin{minipage}[t]{0.32\linewidth}
    \centering
    \parbox[t][\archcompareimageheight][t]{\linewidth}{%
      \centering
      \vspace{0pt}
      \includegraphics[width=\linewidth]{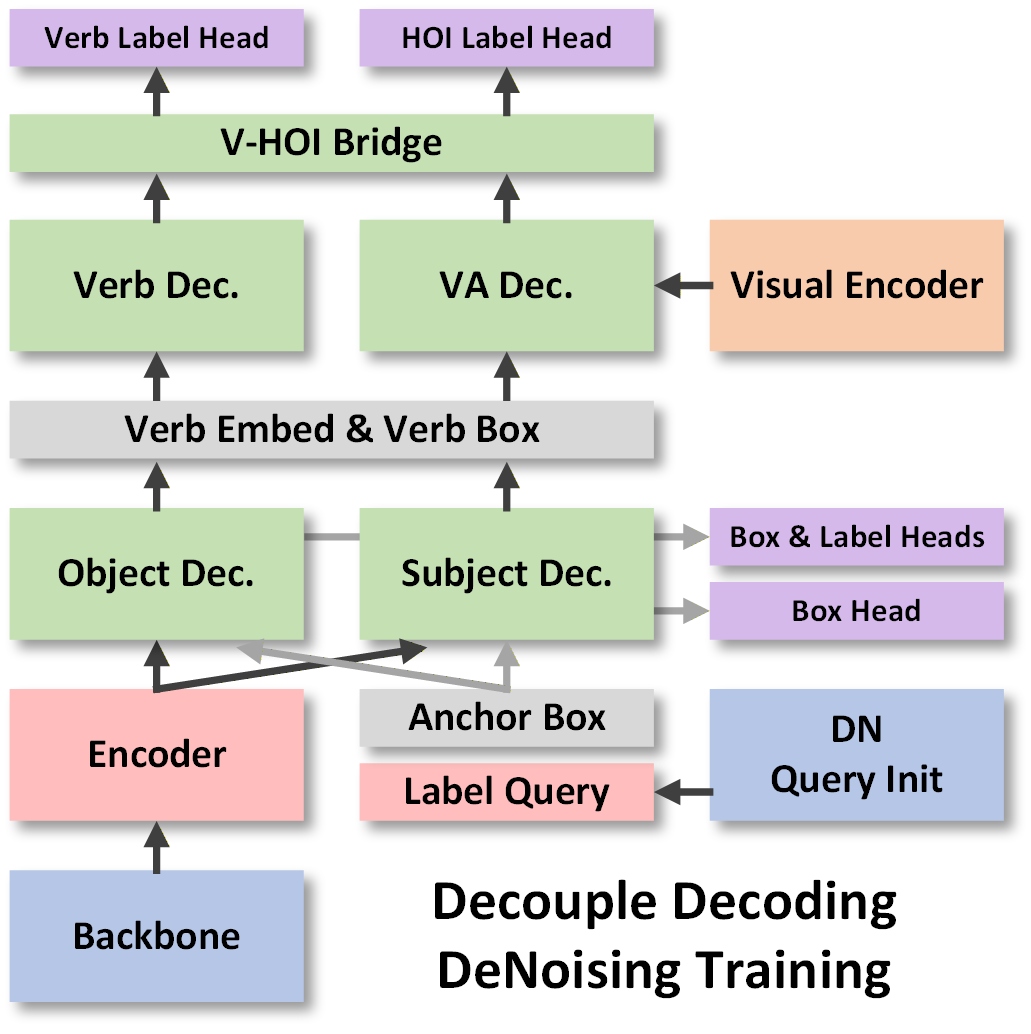}%
    }
    \vspace{2pt}
    {\small\textbf{(a) SOV-STG-VLA}}
  \end{minipage}
  \hfill
  \begin{minipage}[t]{0.32\linewidth}
    \centering
    \parbox[t][\archcompareimageheight][t]{\linewidth}{%
      \centering
      \vspace{0pt}
      \includegraphics[width=\linewidth]{img/arch_compare/Hybrid-SOV-Arch.jpg}%
    }
    \vspace{2pt}
    {\small\textbf{(b) Hybrid-SOV}}
  \end{minipage}
  \hfill
  \begin{minipage}[t]{0.32\linewidth}
    \centering
    \parbox[t][\archcompareimageheight][t]{\linewidth}{%
      \centering
      \vspace{0pt}
      \includegraphics[width=\linewidth]{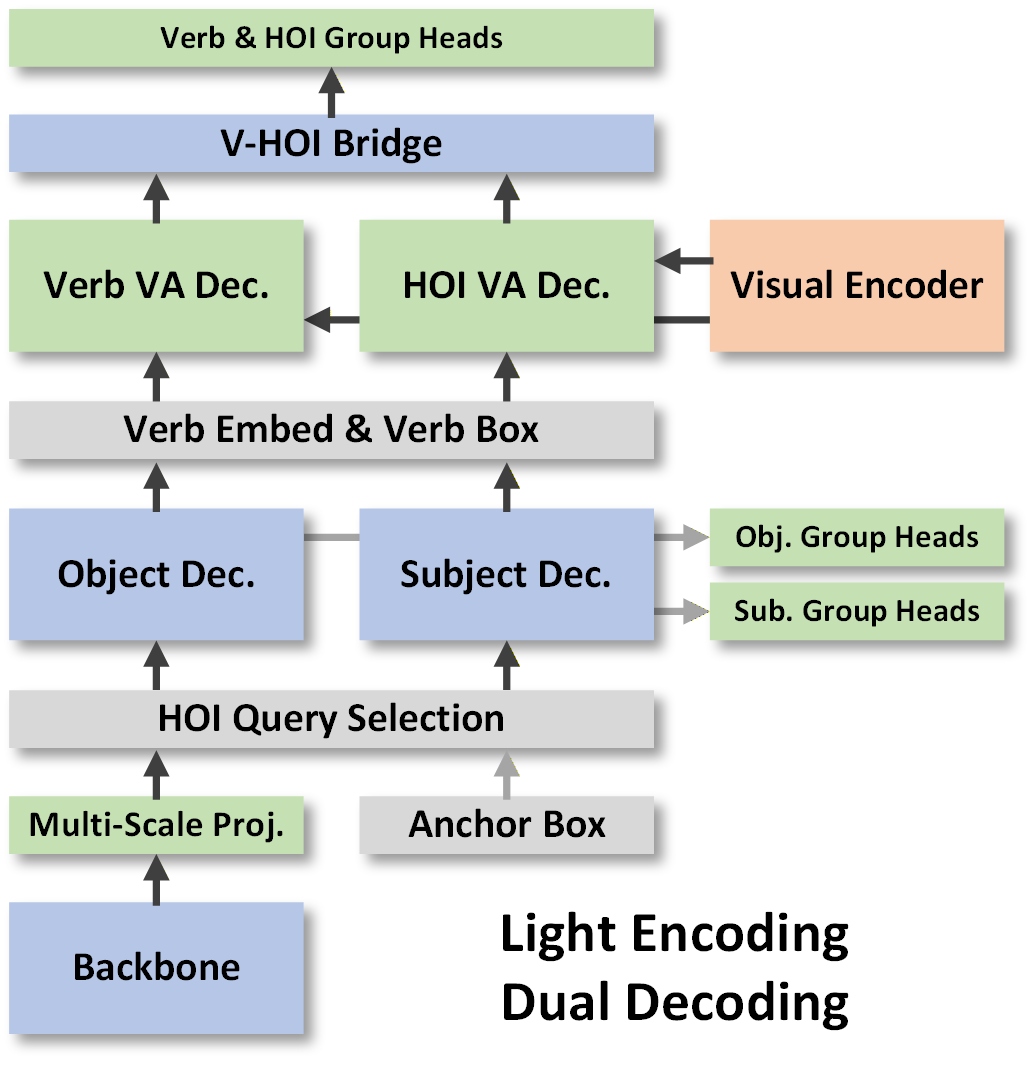}%
    }
    \vspace{2pt}
    {\small\textbf{(c) HOIBlender}}
  \end{minipage}
  \caption{Architectural comparison among SOV-STG-VLA, Hybrid-SOV, and HOIBlender. SOV-STG-VLA uses a Deformable-DETR encoder with target-guided denoising; Hybrid-SOV adopts an RT-DETR-style hybrid encoder and HOI query selection. HOIBlender removes the encoder stage and blends candidates from the multi-scale projector with spatial-semantic decoding and BLIP-2 priors.}
  \label{fig:arch_compare}
\end{figure}

These gaps motivated our prior SOV decoupling line. SOV-STG-VLA~\cite{chen2025focusing} introduced a Subject-Object-Verb decoder with target-guided denoising and a BLIP-2-based Vision-Language Advisor (VLA) that supplies semantic memory inside the verb decoder. Hybrid-SOV~\cite{11367687} further replaced the ResNet~\cite{he2016deep} backbone with a DINOv2-pretrained ViT~\cite{oquabdinov2}, adopted an RT-DETR-style hybrid encoder~\cite{zhao2024detrs}, and constructed HOI queries from coarse predictions over visual features. Figure~\ref{fig:arch_compare} contrasts these predecessors with HOIBlender: SOV-STG-VLA relies on a Deformable-DETR encoder with target-guided denoising; Hybrid-SOV adopts an RT-DETR-style hybrid encoder; HOIBlender removes the encoder stage altogether. This progression suggests two lessons: \textbf{a stronger, content-aware detector foundation can be more valuable than an increasingly heavy interaction head}, and \textbf{vision-language priors are most useful when they shape the predictor's decision space rather than merely calibrating its output}. Yet Hybrid-SOV still spends substantial computation on a dedicated encoder, and its denoising-query schedule becomes cumbersome when scaling the model family.

Building on these observations, we propose \textbf{HOIBlender}. The name is intentional: instead of stacking a detector, an interaction head, and a semantic calibration module as loosely connected components, HOIBlender treats HOI detection as a blending problem. It blends \emph{detector-grounded tokens} from a multi-scale DINOv2 projector, \emph{spatial interaction states} from coupled subject-object decoding, and \emph{semantic priors} from BLIP-2 image and text embeddings. This formulation is motivated by the structure of HOI itself: reliable predictions require visual evidence for the entities, geometry for their relation, and language priors for fine-grained and rare actions to meet inside the same prediction path.

\begin{wrapfigure}{r}{0.5\linewidth}
  \vspace{-0.6em}
  \centering
  \includegraphics[width=\linewidth]{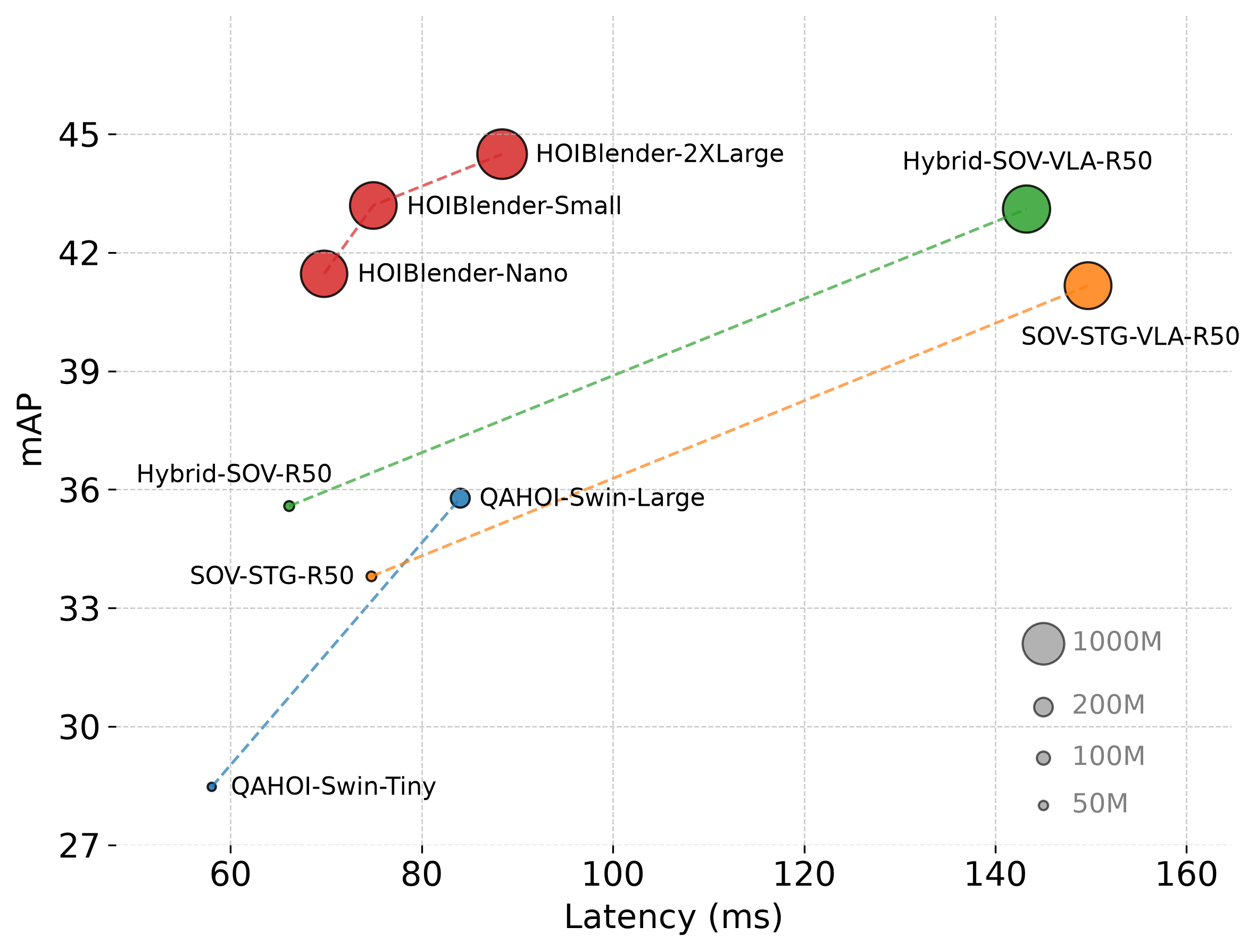}
  \caption{Accuracy-latency-parameter trade-off on HICO-DET. Bubble area encodes the total parameter count; HOIBlender variants establish a favorable frontier across model scales.}
  \label{fig:efficiency_bubble}
\end{wrapfigure}
Concretely, HOIBlender adopts an RF-DETR/LW-DETR-style foundation~\cite{robinson2025rf,chen2024lw} with a DINOv2 backbone~\cite{oquabdinov2} and a multi-scale projector. Rather than refining all tokens with another transformer encoder, it selects top-$K$ image-conditioned tokens directly from the projector as subject and object candidates. The selected states enter a dual-stage decoder: \emph{Stage~A} refines human-object geometry through two coupled streams, while \emph{Stage~B} performs verb and HOI classification with progressive vision-language prior fusion. BLIP-2 features serve as semantic memory, and BLIP-2 text embeddings initialize the verb and HOI classifier weights. To stabilize optimization without increasing inference cost, training employs Group-DETR-style grouped query heads~\cite{chen2023group} that collapse to a single group at test time. Three scales (Nano, Small, 2XL) follow the RF-DETR scaling recipe, yielding a uniform architecture across deployment budgets.

This design addresses the two gaps above in a single pipeline. Removing the encoder eliminates the most expensive structural component while preserving multi-scale visual evidence from a foundation backbone. Fusing vision-language priors at \emph{both} the decoder-memory level and the classifier-weight level gives rare HOI categories a semantically meaningful initialization and a semantically guided decoding path, which is crucial on HICO-DET where many interaction classes have fewer than ten training instances~\cite{chao2018learning}. As previewed in Fig.~\ref{fig:efficiency_bubble}, HOIBlender variants form a favorable accuracy-latency-parameter frontier across three model scales: the Nano variant offers a compact parameter footprint, while the 2XL variant pushes the accuracy frontier. A detailed analysis is provided in Sec.~\ref{sec:efficiency}.

Our contributions are summarized as follows. (1) We propose HOIBlender, an encoder-free HOI detector that selects interaction candidates directly from the multi-scale projector outputs of a DINOv2-based RF-DETR/LW-DETR-style foundation. (2) We design a dual-stage interaction decoder that blends spatial and semantic reasoning: Stage~A stabilizes human-object geometry, and Stage~B performs verb and HOI classification with BLIP-2 features as semantic memory and BLIP-2 text embeddings as classifier initialization. (3) We incorporate grouped-query training to improve optimization while preserving the inference cost of a compact single-group decoder. (4) On HICO-DET, HOIBlender consistently outperforms strong recent baselines including SOV-STG-VLA and Hybrid-SOV-VLA across three model scales, achieving $44.49$ Default Full mAP in only $9$ training epochs with a favorable accuracy-latency-parameter trade-off.

\section{Related Work}

\textbf{From multi-stream to query-based HOI detection.}
Early HOI detectors followed a detect-then-classify paradigm built on top of CNN object detectors~\cite{gkioxari2018detecting,gao2018ican,qi2018learning,wan2019pose,gupta2019no}. They typically combined human, object, and pairwise relation streams, using graph neural networks~\cite{qi2018learning} or attention-based context aggregation~\cite{gao2018ican} to fuse cues. Although these designs established the basic vocabulary for HOI reasoning, they suffered from heavy pair enumeration and limited end-to-end optimization. The DETR family~\cite{carion2020end,zhu2020deformable} replaced this pipeline with set prediction, leading to query-based HOI detectors~\cite{tamura2021qpic,kim2021hotr,zou2021end,chen2021reformulating,Zhang_2022_CVPR}. Later work decomposed and structured the queries: PQNet~\cite{chen2022parallel} introduced parallel detection decoders and a verb decoder with branch-specific queries, QAHOI~\cite{chen2023qahoi} added query-based spatial anchors over multi-scale deformable features, MSTR~\cite{Kim_2022_CVPR} explored multi-scale context attention, and PViC~\cite{zhang2023pvic} introduced box-pair positional embeddings on top of an H-DETR backbone. These methods improve spatial modeling and association reasoning, but most remain tied to vanilla DETR or Deformable DETR foundations and do not fully exploit recent progress in efficient detection.

\textbf{Modern detection architectures and their adoption in HOI.}
The detection community has continued to evolve the DETR paradigm at a fast pace. DAB-DETR~\cite{liu2022dabdetr} introduced dynamic anchor boxes; DN-DETR~\cite{Li_2022_CVPR} and DINO~\cite{Li_2022_CVPR} added denoising training; RT-DETR~\cite{zhao2024detrs} proposed a hybrid encoder that decouples intra-scale and cross-scale interactions; and RF-DETR~\cite{robinson2025rf} together with LW-DETR~\cite{chen2024lw} pushed the efficiency frontier through lightweight architectures, refined training recipes, and ViT-based foundation features. In parallel, self-supervised foundation models such as DINOv2~\cite{oquabdinov2} provide transferable visual features beyond conventional supervised ImageNet pretraining. HOI detection has only recently started to absorb these advances. Hybrid-SOV~\cite{11367687} brought an RT-DETR-style hybrid encoder and a DINOv2 ViT backbone into HOI detection, showing that data-driven HOI query selection from coarse interaction predictions outperforms learned content embeddings. HOIBlender continues this detector-first direction but removes the encoder stage: a DINOv2 backbone and multi-scale projector directly produce the candidate tokens used as subject and object queries.

\textbf{Vision-language priors for HOI understanding.}
Vision-language pretraining has become a core ingredient for long-tail and open-vocabulary HOI~\cite{radford2021learning,li2022blip,li2023blip}. CLIP-based methods~\cite{liao2022gen,ning2023hoiclip,mao2024clip4hoi} initialize the interaction classifier with CLIP text embeddings, distill CLIP image embeddings into HOI features, or build zero-shot retrieval branches on top of CLIP. RLIP and RLIPv2~\cite{yuan2022rlip,yuan2023rlipv2} explicitly pretrain visual-relational representations against language descriptions. UniHOI~\cite{cao2024detecting} aligns human-object pair features with BLIP-2 relation embeddings through prompt-guided decoding, while HOLa~\cite{hola2025iccv}, SICHOI~\cite{luo2024discovering}, and SOV-STG-VLA~\cite{chen2025focusing} explore VLM-guided decoders for verb and HOI classification. A common limitation is that VLM information is often used as a side branch or post-hoc calibration; the visual encoder and interaction decoder still operate largely independently of the VLM features. HOIBlender instead blends language priors into the predictor at two levels: BLIP-2 features are injected as decoder memory, and BLIP-2 text embeddings initialize both the verb and HOI classifiers. This integrates semantic structure into the decision boundary itself rather than applying it as an output bias.

\textbf{Efficient training and grouped queries.}
Group-DETR~\cite{chen2023group} showed that multiple query groups during training enrich matching diversity and accelerate convergence, while a single group is sufficient at inference. This idea is orthogonal to architectural design and is well suited to HOI detection, where long-tail labels make one-to-one matching especially sensitive. HOIBlender adapts grouped-query training to HOI-specific matching and losses, retaining only one group at inference time so that runtime remains close to a compact detector-decoder pipeline.

\begin{figure}[t]
  \centering
  \includegraphics[width=\linewidth]{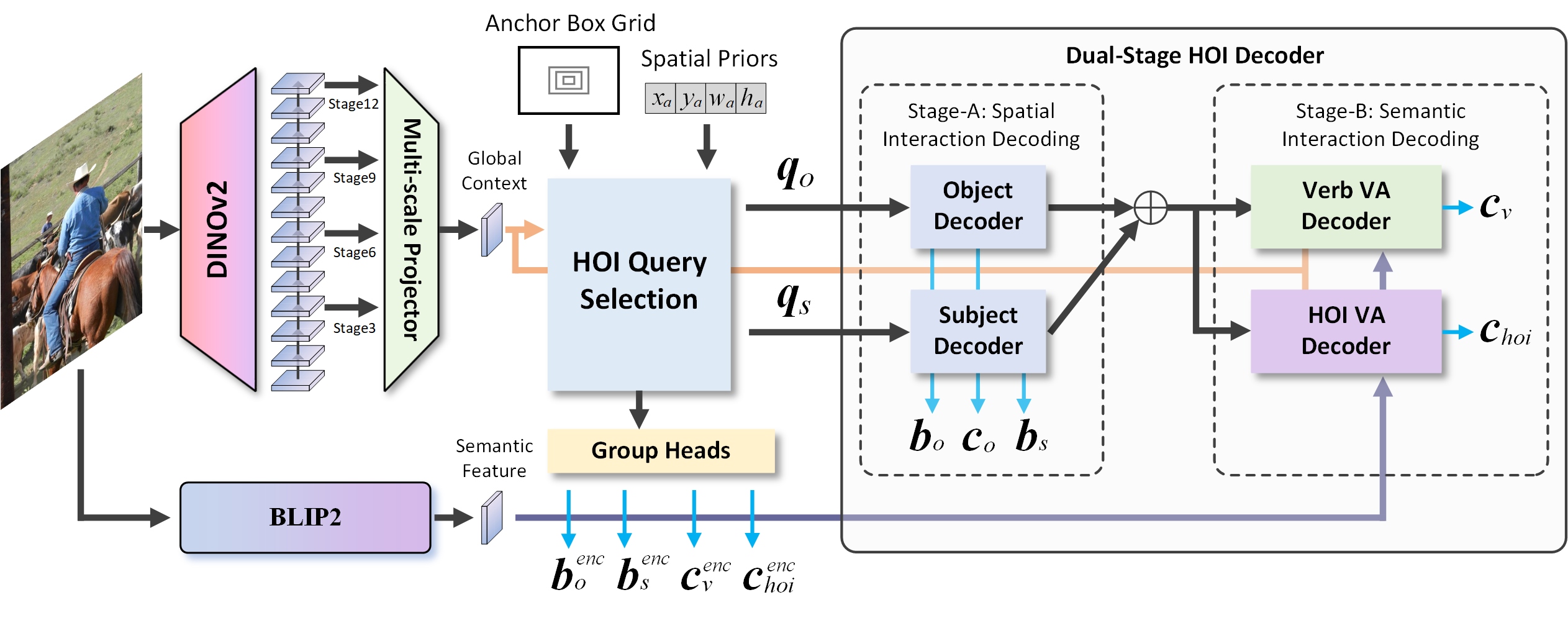}
  \caption{Overall architecture of HOIBlender. A DINOv2 backbone produces multi-scale features that are aggregated by a multi-scale projector. HOIBlender selects top-$K$ image-conditioned tokens as subject and object candidates and feeds them into a dual-stage decoder. Stage~A refines human-object geometry; Stage~B blends verb and HOI decoding with progressive BLIP-2 prior fusion. Grouped query heads are used during training and reduced to a single group at inference time.}
  \label{fig:method_overview}
\end{figure}

\section{Method}

\subsection{Problem Setup and Design Principles}
Given an image $I$, HOI detection predicts a set of interaction instances $(b_h, b_o, c_o, c_{hoi})$, where $b_h$ and $b_o$ denote human and object boxes, $c_o$ denotes the object class, and $c_{hoi}$ denotes an HOI category, equivalently a verb-object composition~\cite{chao2018learning}. HOIBlender jointly outputs object logits, object boxes, subject boxes, verb logits, and HOI logits, coupling geometric grounding and semantic interaction reasoning in a single end-to-end model.

The framework is organized around three design principles distilled from the SOV-STG-VLA~\cite{chen2025focusing} and Hybrid-SOV~\cite{11367687} line. \textbf{(1) Detector-first:} once the visual backbone is strong enough, HOI reasoning should reuse projected detector features directly instead of repeatedly refining them with a dedicated transformer encoder. \textbf{(2) Spatial-then-semantic:} interaction prediction is more stable when human-object geometry is decoded before semantic priors are fused, especially under long-tail supervision. \textbf{(3) Train-rich, infer-compact:} optimization benefits from richer query diversity, but the inference budget should remain compact for deployment. These principles define the operational meaning of ``blending'' in HOIBlender: each source of information is introduced at the stage where it is most reliable. They guide the architecture in Fig.~\ref{fig:method_overview} and the dual-stage decoder in Sec.~\ref{sec:dualstage}.

\subsection{Detector-First Foundation}
\label{sec:detector_first}

HOIBlender uses a DINOv2 ViT backbone~\cite{oquabdinov2} together with a multi-scale projector, following the structural design of RF-DETR/LW-DETR~\cite{robinson2025rf,chen2024lw}. The backbone produces multi-scale visual features, and the projector aggregates them into a sequence of context-rich tokens. Unlike earlier DETR-style HOI detectors that refine all feature tokens with a transformer encoder before decoding~\cite{tamura2021qpic,zhang2023pvic,11367687}, HOIBlender feeds the projected tokens directly into the HOI decoder. As shown in Fig.~\ref{fig:arch_compare}, this is a stricter detector-first design than Hybrid-SOV, which still places an encoder between the backbone and the HOI decoders.

\textbf{Direct HOI query selection.} A query selection module operates on the projected token sequence and predicts an objectness score and an interaction-relevance score for each token. The top-$K$ tokens form two parallel sets used as subject queries and object queries. Because these queries are constructed from the multi-scale projector outputs of a strong self-supervised backbone, they are conditioned on the input image and provide a content-aware initialization for the dual-stage decoder. This refines the HOI query selection idea introduced by Hybrid-SOV~\cite{11367687}: instead of selecting candidates from the output of an additional encoder, HOIBlender selects them directly from projected backbone features, removing the encoder from the inference path while preserving high-quality initialization.

\textbf{Grouped-query training.} During training we instantiate the prediction head as $G$ parallel query groups in the spirit of Group-DETR~\cite{chen2023group}. Each group is matched to ground-truth HOI triplets independently with a separate Hungarian assignment, different groups can specialize to different assignments, enriching optimization signals without changing the inference architecture. 
At inference time only one group is kept, so runtime is the same as a compact single-group decoder. This implements the train-rich, infer-compact principle: query diversity is concentrated in training, while deployment latency is determined by a single group.

\subsection{Dual-Stage HOI Decoding}
\label{sec:dualstage}

\begin{wrapfigure}{r}{0.49\linewidth}
  \centering
  \includegraphics[width=\linewidth]{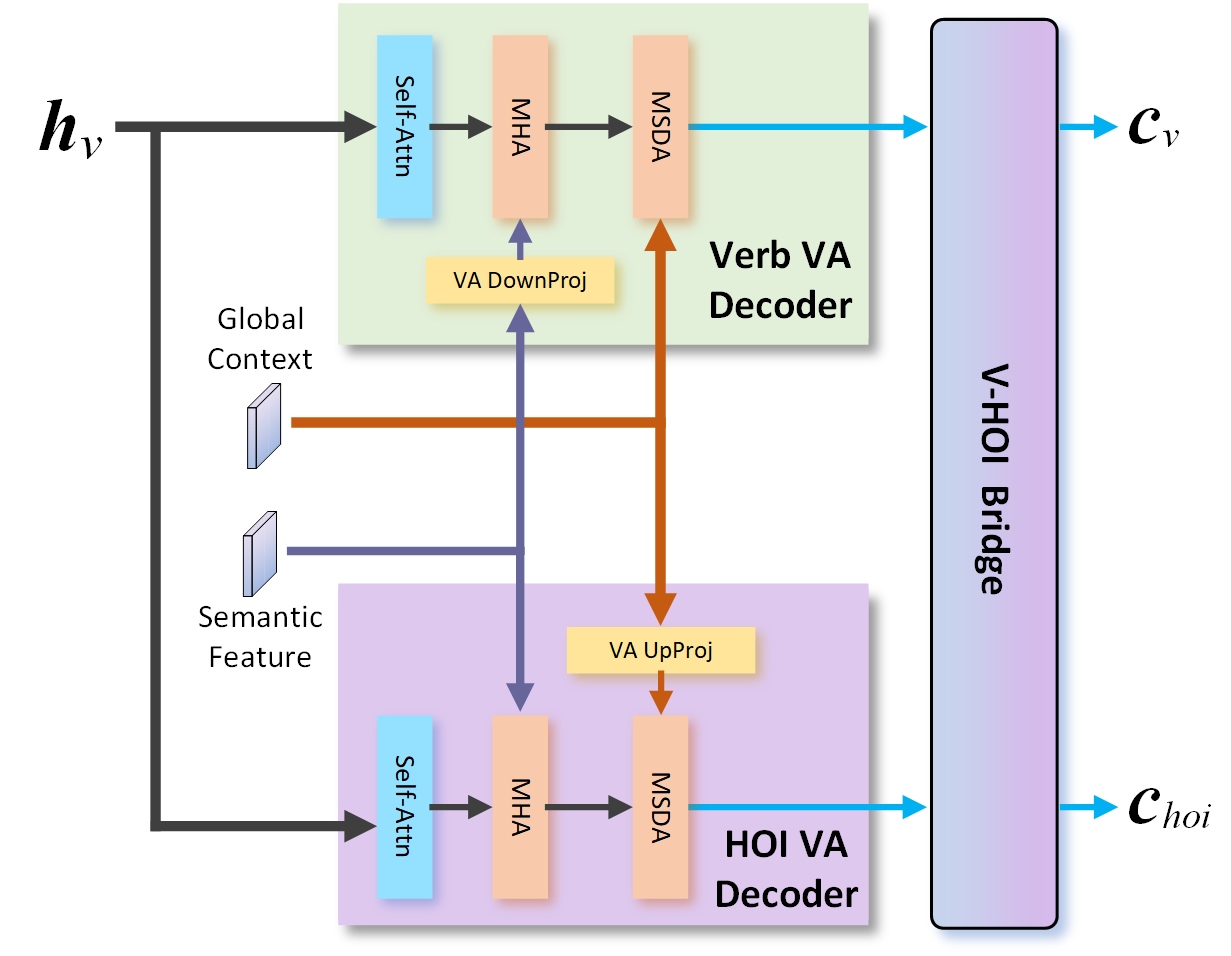}
  \caption{Stage~B of HOIBlender. Verb and HOI vision-advisor decoders consume BLIP-2 memory, and the V-HOI Bridge fuses verb-aware and HOI-aware embeddings.}
  \label{fig:dual_stage_decoder}
\end{wrapfigure}
The HOI decoder is organized as a two-stage interaction reasoning process. Stage~A focuses on spatial geometry, and Stage~B focuses on semantic interaction. This decomposition reflects the natural structure of HOI prediction: a model should first establish where the interacting human and object are, and then decide how they interact. Decoupling these roles reduces the risk that noisy semantic activations disturb early spatial refinement, which is a known weakness of single-stage interaction heads~\cite{chen2025focusing}.

\textbf{Stage~A: spatial interaction decoding.}
Starting from the selected subject and object queries, Stage~A uses two coupled decoder streams to refine interaction geometry. One stream predicts subject-aware states and human boxes, while the other predicts object-aware states and object boxes. The streams share the same proposal origin and decoder depth, and exchange information through cross-attention to the projected backbone features. Because both streams originate from the same selected detector tokens, the model can maintain human-object alignment without explicitly enumerating all possible pairs. Let $q_s^l$ and $q_o^l$ denote subject and object query states at decoder layer $l$. After $L$ decoder layers, Stage~A outputs final spatial states $q_s^L, q_o^L$ together with subject and object box predictions $\hat{b}_s$ and $\hat{b}_o$. These states are passed to Stage~B for semantic decoding.

\textbf{Stage~B: semantic decoding with progressive BLIP-2 priors.}
Stage~B performs verb and HOI classification with explicit vision-language guidance. As illustrated in Fig.~\ref{fig:dual_stage_decoder}, both branches are implemented as vision-advisor decoders~\cite{chen2025focusing,11367687} that consume BLIP-2 features~\cite{li2023blip} as additional memory through a linear dimension-reduction layer. In contrast to SOV-STG-VLA and Hybrid-SOV, which apply the vision advisor only to the HOI branch, HOIBlender extends it to the verb branch, providing finer-grained semantic guidance for predicate prediction. The V-HOI Bridge module~\cite{chen2025focusing} then fuses verb-aware and HOI-aware embeddings, encouraging the HOI branch to reuse verb-specialized evidence rather than reconstructing it independently.

We aggregate spatial states into an interaction representation $h_v$ that drives the verb branch, and let the HOI branch produce a representation $h_{hoi}$ that is normalized and projected against the BLIP-2-initialized HOI classifier:
\begin{equation}
h_v = \frac{q_s^L + q_o^L}{2}, \qquad
z_{hoi} = s \cdot W_{hoi}\!\left(\frac{h_{hoi}}{\|h_{hoi}\|_2}\right),
\end{equation}
where $s$ is a learnable temperature that stabilizes similarity-based prediction. Verb logits are produced through an analogous projection-normalization-temperature pipeline. By placing prior fusion after Stage~A, the decoder ensures that semantic priors interact with stabilized human-object geometry rather than noisy spatial states, which empirically benefits long-tail HOI categories.

\subsection{Multi-Prior Semantic Fusion}
\label{sec:fusion}

HOIBlender injects vision-language priors at two complementary levels. \textbf{Image-conditioned priors:} BLIP-2 features are fed as memory to the vision-advisor decoders for both the verb and HOI branches, providing global semantic context that complements the local detector features used in cross-attention to the multi-scale projector. \textbf{Label-conditioned priors:} HOI text prompts and verb text prompts are encoded once with a frozen BLIP-2 text encoder and used to initialize the corresponding classifier weights $W_{verb}$ and $W_{hoi}$.

The two prior paths are deliberately separated. Image-conditioned BLIP-2 memory provides scene-level evidence, such as the global object context and affordance cues that may be weak in a local human-object crop, and is consumed dynamically by the verb and HOI advisor decoders for each input image. Label-conditioned text embeddings, in contrast, are fixed before training and define the geometry of the output label space. This separation prevents the semantic prior from acting only as a late calibration term: the decoder can attend to image-specific semantic memory while the classifier begins from a label space in which related verbs and verb-object compositions are already close.

The label-conditioned initialization is particularly valuable for rare HOI categories. When a category appears only a few times during training, gradient supervision alone is too sparse to position its classifier weight reliably, and random initialization places semantically related labels arbitrarily in the embedding space. BLIP-2 text embeddings provide a structured HOI label geometry, giving the classifier a semantic organization from the first training step. The detector can therefore adapt visual features to an already meaningful label space instead of constructing that space from sparse supervision alone. This is a key reason HOIBlender remains competitive on the long-tailed Rare split despite using a lighter detector-centric pipeline.

\subsection{Matching and Training Objectives}

We train HOIBlender end-to-end with Hungarian matching~\cite{kuhn1955hungarian}. The matching cost combines object classification, subject and object box regression, generalized IoU~\cite{rezatofighi2019generalized}, and interaction-related classification terms. The overall objective is
\begin{equation}
\begin{array}{l}
\mathcal{L} = \lambda_{vfl}\mathcal{L}_{vfl} + \lambda_{box}(\mathcal{L}^{box}_{sub} + \mathcal{L}^{box}_{obj}) \\[1pt]
\quad + \lambda_{giou}(\mathcal{L}^{giou}_{sub} + \mathcal{L}^{giou}_{obj}) + \lambda_{verb}\mathcal{L}_{verb} + \lambda_{hoi}\mathcal{L}_{hoi},
\end{array}
\end{equation}
where $\mathcal{L}_{vfl}$ is the object-focused varifocal-style classification loss~\cite{zhao2024detrs}, $\mathcal{L}_{verb}$ and $\mathcal{L}_{hoi}$ are multi-label interaction losses, and the box/GIoU terms supervise human and object localization. Auxiliary supervision is applied to all intermediate decoder layers to stabilize training. In practice, AdamW~\cite{loshchilov2018decoupled}, mixed precision, gradient accumulation, and grouped-query training~\cite{chen2023group} are jointly used for efficient optimization. Because all loss terms operate on one end-to-end pipeline, detector efficiency, dual-stage decoding, and semantic priors are optimized together rather than as separately tuned modules.

\section{Experiments}

\subsection{Dataset and Evaluation Protocol}
We evaluate HOIBlender on HICO-DET~\cite{chao2018learning}, a widely used HOI detection benchmark containing $38{,}118$ training and $9{,}658$ test images annotated with $80$ object categories, $117$ verbs, and $600$ HOI categories. Following standard protocols, we report mean Average Precision (mAP) under both the \textit{Default} and \textit{Known Object} settings, each split into \textit{Full}, \textit{Rare}, and \textit{Non-Rare} subsets. The \textit{Rare} subset contains the $138$ HOI categories with fewer than $10$ training instances, making it a direct test of tail-category robustness.

\subsection{Implementation Details}
We instantiate three HOIBlender variants following the RF-DETR scaling strategy. The pretrained backbone input resolution is $384$, $512$, and $880$ for HOIBlender-N, HOIBlender-S, and HOIBlender-2XL, respectively, and the corresponding decoder depth is $2$, $3$, and $5$. Compared with Hybrid-SOV, which uses six decoder layers after an RT-DETR-style encoder, this reduces depth while preserving accuracy. All variants use $13$ grouped query heads during training and one group at inference time. Training uses $15$ epochs for the Nano and Small models and $9$ epochs for the 2XL model; the learning-rate drop is set to epoch $13$ for the $15$-epoch schedule and epoch $6$ for the 2XL schedule. Optimization uses AdamW~\cite{loshchilov2018decoupled} with learning rate $1\times10^{-4}$, encoder learning rate $1.5\times10^{-4}$, weight decay $1\times10^{-4}$, mixed precision, and gradient accumulation. The training is on 8 NVIDIA A6000 GPUs with batch size 16, and latency is measured on a single NVIDIA A100 GPU with batch size $1$, averaged over $100$ samples on the HICO-DET test set.

\begin{table*}[t]
  \caption{Comparison on HICO-DET. HOIBlender improves over recent SOV-style baselines across three model scales while preserving an efficient training schedule.}
  \label{tab:sota_comparison_hico}
  \centering
  \resizebox{0.95\textwidth}{!}{
    \begin{tabular}{lcccccccc}
      \toprule
      \multirow{2}{*}{Method} & \multirow{2}{*}{Epoch} & \multirow{2}{*}{Backbone} & \multicolumn{3}{c}{Default} & \multicolumn{3}{c}{Known Object} \\
      \cmidrule(lr){4-6} \cmidrule(lr){7-9}
      & & & Full & Rare & Non-Rare & Full & Rare & Non-Rare \\
      \midrule
      QPIC~\cite{tamura2021qpic}       & 150 & ResNet-50  & 29.07 & 21.85 & 31.23 & 31.68 & 24.14 & 33.93 \\
      MSTR~\cite{Kim_2022_CVPR}        & 50  & ResNet-50  & 31.17 & 25.31 & 32.92 & 34.02 & 28.83 & 35.57 \\
      UPT~\cite{Zhang_2022_CVPR}       & 20  & ResNet-101 & 32.62 & 28.62 & 33.81 & 36.08 & 31.41 & 37.47 \\ 
      RLIP-ParSe~\cite{yuan2022rlip}   & 90  & ResNet-50  & 32.84 & 34.63 & 26.85 & -     & -     & -     \\
      GEN-VLKT-S~\cite{liao2022gen}    & 90  & ResNet-50  & 33.75 & 29.25 & 35.10 & 36.78 & 32.75 & 37.99 \\
      HOICLIP~\cite{ning2023hoiclip}   & 90  & ResNet-50  & 34.69 & 31.12 & 35.74 & 37.61 & 34.47 & 38.54 \\
      ViPLO~\cite{park2023viplo}       & 8   & ViT-B/32   & 34.95 & 33.83 & 35.28 & 38.15 & 36.77 & 38.56 \\
      GEN-VLKT-L~\cite{liao2022gen}    & 90  & ResNet-101 & 34.96 & 31.18 & 36.08 & 38.22 & 34.36 & 39.37 \\
      CLIP4HOI~\cite{mao2024clip4hoi}  & 100 & ResNet-50  & 35.33 & 33.95 & 35.74 & 37.19 & 35.27 & 37.77 \\
      HOLa~\cite{hola2025iccv}         & 12  & ResNet-50  & 39.05 & 38.66 & 39.17 & -     & -     & -     \\
      UniHOI-S~\cite{cao2024detecting} & 90  & ResNet-50  & 40.06 & 39.91 & 40.11 & 42.20 & 42.60 & 42.08 \\
      UniHOI-L~\cite{cao2024detecting} & 90  & ResNet-101 & 40.95 & 40.27 & 41.32 & 43.26 & 43.12 & 43.25 \\
      SICHOI~\cite{luo2024discovering} & 30  & ResNet-50  & 41.79 & 42.38 & 41.61 & 44.27 & 43.64 & 44.46 \\
      QAHOI~\cite{chen2023qahoi}       & 150 & Swin-Large & 35.78 & 29.80 & 37.56 & 37.59 & 31.36 & 39.36 \\
      FGAHOI~\cite{ma2023fgahoi}       & 190 & Swin-Large & 37.18 & 30.71 & 39.11 & 38.93 & 31.93 & 41.02 \\
      PViC~\cite{zhang2023pvic}        & 30  & Swin-Large & 44.32 & \textbf{44.61} & 44.24 & \textbf{47.81} & \textbf{48.38} & \textbf{47.64} \\
      \midrule
      SOV-STG-VLA~\cite{chen2025focusing} & 15 & ResNet-50 & 41.16 & 39.48 & 41.67 & 43.81 & 42.63 & 44.17 \\
      Hybrid-SOV-VLA~\cite{11367687} & 15 & ResNet-50 & 43.10 & 43.04 & 43.12 & 46.02 & 46.14 & 45.98 \\
      \midrule
      \textbf{HOIBlender-N} & 15 & DINOv2-S & \oursNanoDefFull & \oursNanoDefRare & \oursNanoDefNonRare & \oursNanoKnownFull & \oursNanoKnownRare & \oursNanoKnownNonRare \\
      \textbf{HOIBlender-S} & 15 & DINOv2-S & \oursSmallDefFull & \oursSmallDefRare & \oursSmallDefNonRare & \oursSmallKnownFull & \oursSmallKnownRare & \oursSmallKnownNonRare \\
      \textbf{HOIBlender-2XL} & 9 & DINOv2-B & \textbf{\oursXXLDefFull} & \oursXXLDefRare & \textbf{\oursXXLDefNonRare} & \oursXXLKnownFull & \oursXXLKnownRare & \oursXXLKnownNonRare \\
      \bottomrule
    \end{tabular}
  }
\end{table*}

\subsection{Comparison with the State of the Art}
Table~\ref{tab:sota_comparison_hico} compares HOIBlender with representative HOI detectors using ResNet~\cite{he2016deep}, Swin Transformer~\cite{liu2021swin}, and ViT-based~\cite{dosovitskiy2021an} backbones. HOIBlender-N reaches $\oursNanoDefFull$ Default Full mAP and $\oursNanoDefRare$ Rare mAP with only $15$ training epochs and a DINOv2-S backbone, surpassing SOV-STG-VLA~\cite{chen2025focusing} on every split. HOIBlender-S improves to $\oursSmallDefFull$ Default Full mAP, matching or exceeding Hybrid-SOV-VLA~\cite{11367687} while using fewer decoder layers and no transformer encoder. The largest variant, HOIBlender-2XL, achieves $\oursXXLDefFull$ Default Full and $\oursXXLDefNonRare$ Non-Rare mAP using only $9$ training epochs, the best Default Full result among the methods listed. These results show that the detector-first blend scales consistently across model sizes. PViC~\cite{zhang2023pvic} remains strong on the Rare split and some Known Object metrics, indicating that detector-side improvements alone do not fully solve long-tail HOI; this motivates the semantic-prior fusion analyzed below.

\begin{table*}[t]
  \begin{minipage}{0.64\linewidth}
    \centering
    \caption{Component ablations under the Default protocol, comparing Hybrid-SOV-VLA-R50 (left column in each split) and HOIBlender-2XL (right column in each split). * indicates results reproduced by our implementation.}
    \label{tab:component_contribution}
    \resizebox{\linewidth}{!}{
      \begin{tabular}{@{}lcccccc@{}}
        \hline
        \multicolumn{1}{l|}{\multirow{2}{*}{Method}} & \multicolumn{6}{c}{Default (Hybrid-SOV vs. HOIBlender)}\\
        \multicolumn{1}{l|}{}                     & \multicolumn{2}{c}{\textit{Full}} & \multicolumn{2}{c}{\textit{Rare}} & \multicolumn{2}{c}{\textit{Non-Rare}} \\ \hline \hline
        \multicolumn{1}{l|}{baseline}            & \multicolumn{1}{c}{43.10} & \multicolumn{1}{c|}{\textbf{44.49}} & \multicolumn{1}{c}{43.04} & \multicolumn{1}{c|}{\textbf{43.21}} & \multicolumn{1}{c}{43.12} & \multicolumn{1}{c}{\textbf{44.86}} \\
        \multicolumn{1}{l|}{w/o Verb Dec.}            & \multicolumn{1}{c}{43.05} & \multicolumn{1}{c|}{43.92} & \multicolumn{1}{c}{42.09} & \multicolumn{1}{c|}{42.11} & \multicolumn{1}{c}{43.33} & \multicolumn{1}{c}{44.46} \\
        \multicolumn{1}{l|}{w/o Verb Dec. \& w/o VLA}     & \multicolumn{1}{c}{32.36*} & \multicolumn{1}{c|}{34.26} & \multicolumn{1}{c}{23.92*} & \multicolumn{1}{c|}{24.69} & \multicolumn{1}{c}{34.88*} & \multicolumn{1}{c}{37.12}  \\
        \multicolumn{1}{l|}{w/o Subject Dec. \& w/o VLA}     & \multicolumn{1}{c}{35.50} & \multicolumn{1}{c|}{36.97} & \multicolumn{1}{c}{30.59} & \multicolumn{1}{c|}{30.65} & \multicolumn{1}{c}{36.96} & \multicolumn{1}{c}{38.85}  \\ \hline
      \end{tabular}
    }
  \end{minipage}
  \hfill
  \begin{minipage}{0.34\linewidth}
    \centering
    \caption{Ablation on the number of grouped heads for HOIBlender-N. ``1'' denotes training without any additional grouped heads.}
    \label{tab:ablation_group_num}
    \resizebox{\linewidth}{!}{
      \begin{tabular}{cccc}
        \toprule
        Group Num. & Full & Rare & Non-Rare \\
        \midrule
        13 & \textbf{\oursNanoDefFull} & \textbf{\oursNanoDefRare} & \textbf{\oursNanoDefNonRare} \\
        6 & \ablSixgroupFull & \ablSixgroupRare & \ablSixgroupNonRare \\
        1 & \ablOnegroupFull & \ablOnegroupRare & \ablOnegroupNonRare \\
        \bottomrule
      \end{tabular}
    }
  \end{minipage}
\end{table*}

\subsection{Ablation Study}

\textbf{Module contribution.}
Table~\ref{tab:component_contribution} compares the contribution of each module on Hybrid-SOV-VLA (left) and HOIBlender-2XL (right) under the Default protocol. Removing the verb decoder (\emph{w/o Verb Dec.}) yields a modest Full-split drop for HOIBlender-2XL ($44.49\!\rightarrow\!43.92$, a relative decrease of $1.28\%$), but the Rare split drops more substantially ($43.21\!\rightarrow\!42.11$, a relative decrease of $2.55\%$), confirming that the verb branch is especially helpful for tail interactions. Further removing the BLIP-2 vision-language advisor (\emph{w/o Verb Dec. \& w/o VLA}) collapses Rare mAP to $24.69$, representing a severe relative performance reduction of $42.86\%$ from the baseline, which is the largest degradation in the table. This isolates the role of multi-prior semantic fusion: without BLIP-2 memory and BLIP-2-initialized classifier weights, rare HOI categories lose the semantic anchor that sparse supervision cannot reliably recover. Disabling the subject decoder along with VLA (\emph{w/o Subject Dec. \& w/o VLA}) results in a Default Full mAP of $36.97$ and a Rare mAP of $30.65$, corresponding to relative decreases of $16.90\%$ and $29.07\%$ from the full HOIBlender-2XL model respectively, showing that the dual-stream spatial design is also load-bearing. Moreover, these combined ablation scenarios demonstrate that semantic priors (VLA) and structural interaction components (Verb and Subject Decoders) are highly synergistic; removing them together deprives the model of both geometric precision and semantic grounding, leading to compounded performance penalties. The same pattern is consistent across Hybrid-SOV-VLA and HOIBlender-2XL, indicating that dual-stage decoding and prior fusion remain effective after removing the encoder and upgrading the backbone to DINOv2.

\textbf{Grouped queries.}
Table~\ref{tab:ablation_group_num} examines grouped-query training with HOIBlender-N. Using $13$ groups gives the best result across all splits. Reducing to $6$ groups drops Default Full mAP from $\oursNanoDefFull$ to $\ablSixgroupFull$ (a relative decrease of $1.86\%$), and using a single group during training, i.e., the standard training schedule without group heads, leads to a larger drop to $\ablOnegroupFull$ Full mAP (a relative decrease of $10.11\%$) and $\ablOnegroupRare$ Rare mAP (a relative decrease of $6.24\%$). This validates the train-rich, infer-compact principle: multi-group matching significantly improves optimization and effectively prevents the model from being trapped in local optima under long-tail distributions, while inference cost is unaffected because only one group is retained at test time.

\begin{figure*}[t]
  \centering
  {\setlength{\tabcolsep}{1.2pt}\renewcommand{\arraystretch}{0.92}
  \resizebox{\linewidth}{!}{%
  \begin{tabular}{@{}c@{\hspace{3pt}}ccccc@{}}
    \raisebox{-0.1\height}{\makebox[0.035\textwidth][c]{\rotatebox[origin=c]{90}{\footnotesize\textbf{Hybrid-SOV-VLA}}}} &
    \raisebox{-0.5\height}{\includegraphics[height=0.18\textwidth]{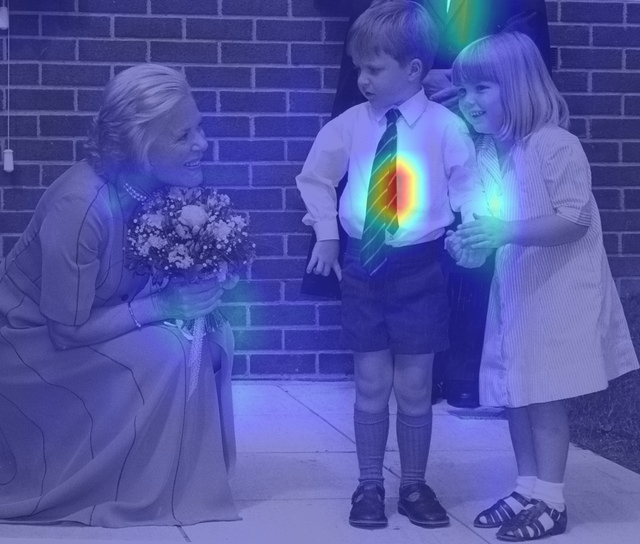}} &
    \raisebox{-0.5\height}{\includegraphics[height=0.18\textwidth]{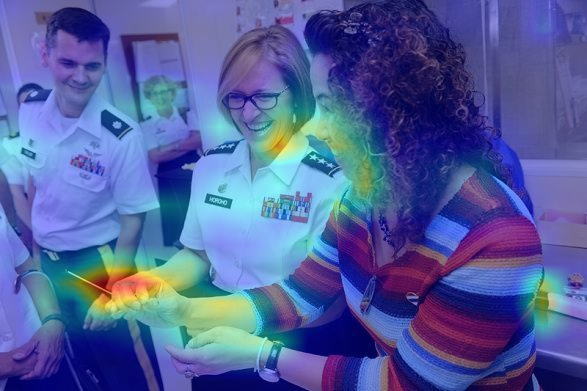}} &
    \raisebox{-0.5\height}{\includegraphics[height=0.18\textwidth]{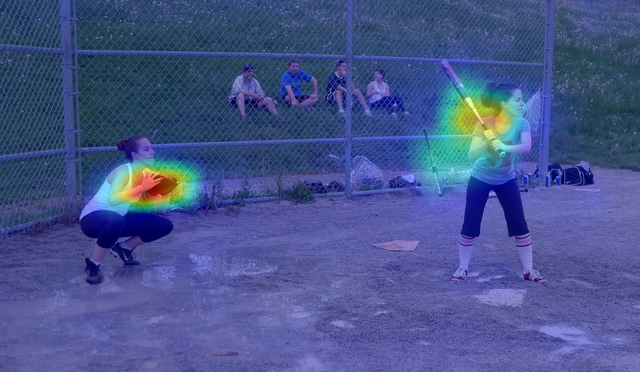}} &
    \raisebox{-0.5\height}{\includegraphics[height=0.18\textwidth]{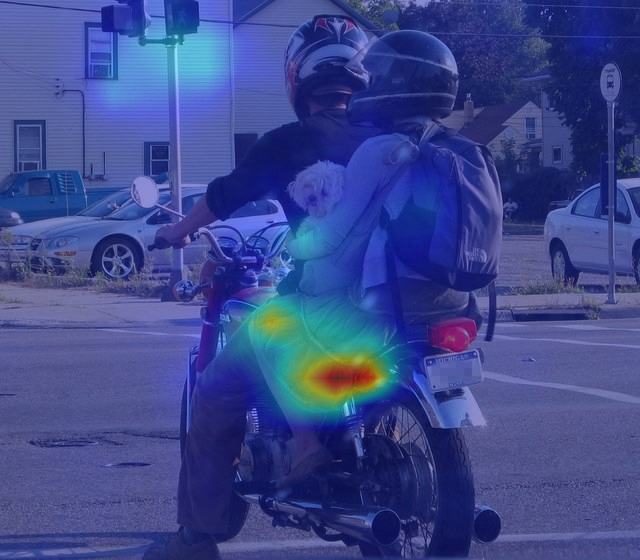}} &
    \raisebox{-0.5\height}{\includegraphics[height=0.18\textwidth]{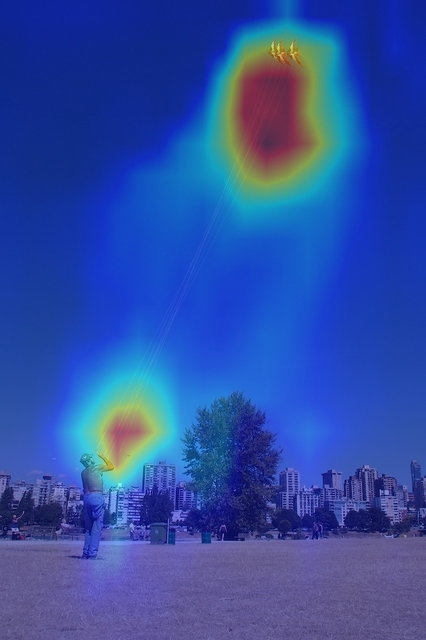}} \\
    \noalign{\vskip 3pt}
    \raisebox{-0.1\height}{\makebox[0.035\textwidth][c]{\rotatebox[origin=c]{90}{\footnotesize\textbf{HOIBlender-2XL}}}} &
    \raisebox{-0.5\height}{\includegraphics[height=0.18\textwidth]{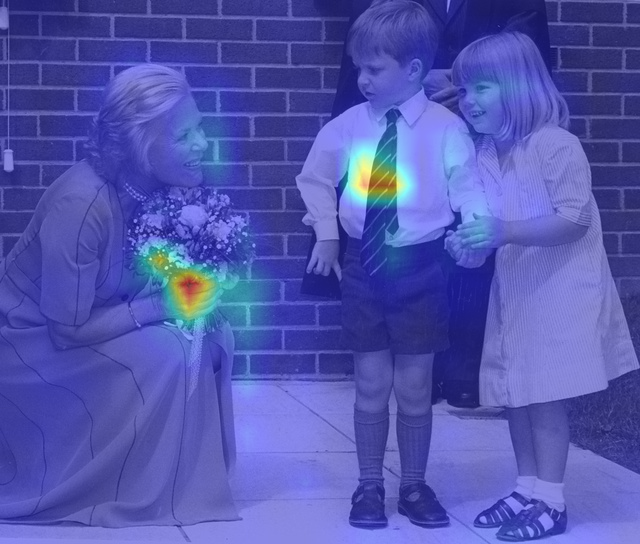}} &
    \raisebox{-0.5\height}{\includegraphics[height=0.18\textwidth]{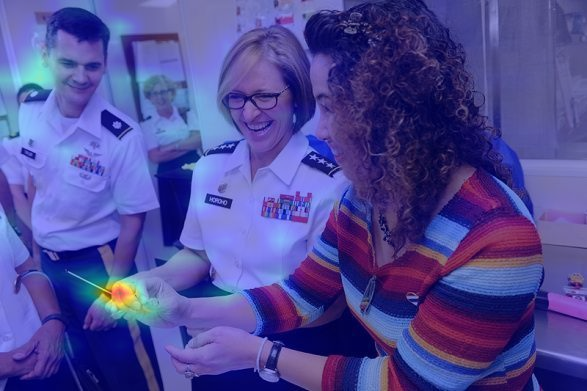}} &
    \raisebox{-0.5\height}{\includegraphics[height=0.18\textwidth]{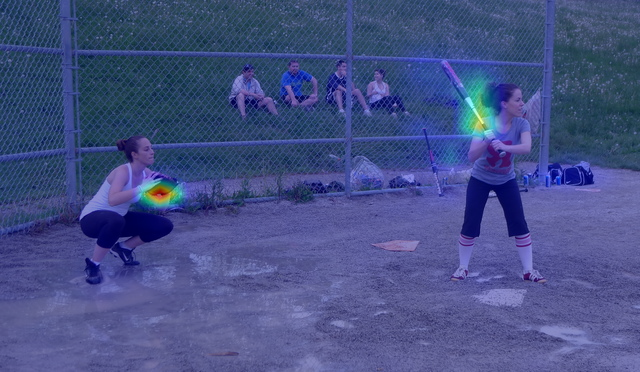}} &
    \raisebox{-0.5\height}{\includegraphics[height=0.18\textwidth]{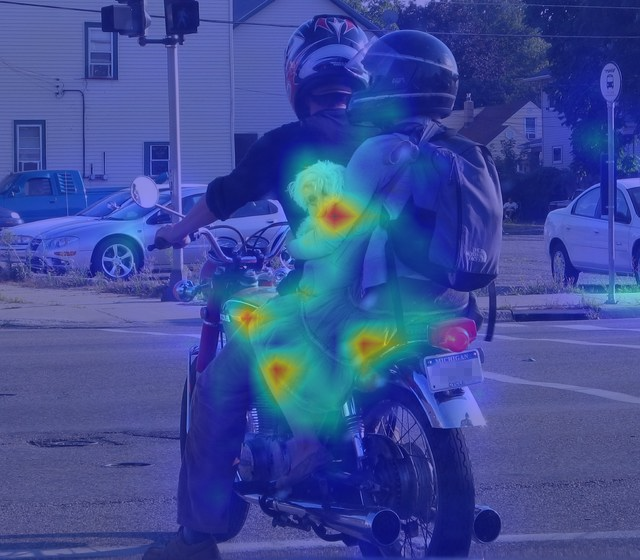}} &
    \raisebox{-0.5\height}{\includegraphics[height=0.18\textwidth]{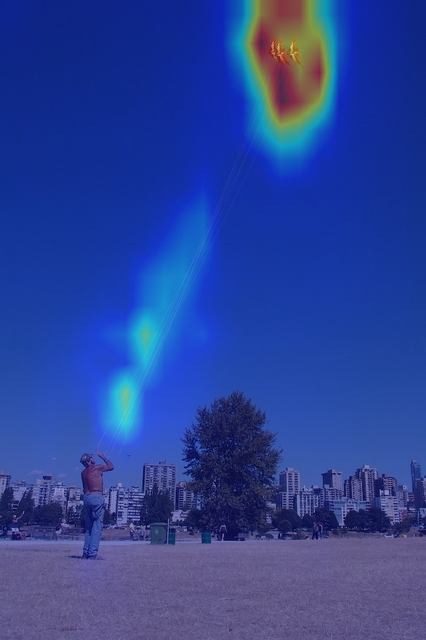}}
  \end{tabular}%
  }
  }
  \caption{Qualitative comparison of predicted verb score maps in HOI query selection. Each column shows the same HICO-DET test image. Warmer colors indicate higher verb scores. HOIBlender-2XL produces more compact activations around human-object interaction evidence than Hybrid-SOV-VLA-R50.}
  \label{fig:score_map_compare}
\end{figure*}

\subsection{Qualitative Analysis}
Figure~\ref{fig:score_map_compare} compares verb score maps produced by the HOI query selection module of Hybrid-SOV and HOIBlender on the same HICO-DET test images.
The score maps from the same level of the multi-scale feature pyramid are visualized.
Hybrid-SOV often produces broad activations that spread across partial object regions or nearby context, diluting the verb-aware information passed to subsequent decoders. HOIBlender produces more compact responses centered on human-object interaction evidence.
For example, in the last column, in HOIBlender, the string of the kite also receives a high verb score, which is a strong cue for the ``fly kite'' interaction, while in Hybrid-SOV the activations are dispersed across the kite and the surrounding area.
This visual pattern aligns with the quantitative gains in Table~\ref{tab:sota_comparison_hico}: by selecting candidates directly from a strong DINOv2 backbone and refining them with the dual-stage decoder, HOIBlender concentrates verb-aware activation on the image regions that carry interaction information. The comparison supports two design principles introduced in Section~\ref{sec:detector_first} and \ref{sec:dualstage}: detector-first query selection provides content-aware candidates, and spatial-then-semantic decoding keeps verb evidence focused rather than diffused across the scene.

\subsection{Efficiency Analysis}
\label{sec:efficiency}

Figure~\ref{fig:efficiency_bubble} visualizes the trade-off among detection quality, inference speed, and model size. We use Default Full mAP and per-image latency in milliseconds as the axes and encode total parameters by bubble area, enabling a direct cross-family comparison. We include SOV-STG-VLA~\cite{chen2025focusing} and Hybrid-SOV-VLA~\cite{11367687} as recent baselines from our line of work. First, HOIBlender-N reaches $\oursNanoDefFull$ mAP at $69.8$ ms per image, providing a compact operating point for resource-constrained settings. Second, HOIBlender-S improves to $\oursSmallDefFull$ mAP at $74.9$ ms, occupying a clear sweet spot: it matches or exceeds Hybrid-SOV-VLA in accuracy with a lighter latency budget, directly reflecting encoder removal and the reduction from six decoder layers to three. Third, HOIBlender-2XL pushes the accuracy frontier to $\oursXXLDefFull$ mAP at $88.4$ ms, indicating that the detector-first pipeline scales gracefully when more capacity is available. These results show that detector efficiency, structured spatial-then-semantic decoding, and deep vision-language integration can be blended within one lightweight pipeline.

\section{Limitations}
BLIP-2 priors improve long-tail recognition, however, the bottleneck of the efficiency also lies in the heavy BLIP-2 vision encoder, which accounts for a large portion of the total latency. Future work could explore more efficient vision-language fusion strategies or lighter vision-language models to further reduce inference cost while preserving the long-tail benefits.

\section{Conclusion}
We presented HOIBlender, an HOI detection framework that blends lightweight detector tokens, spatial interaction decoding, and BLIP-2 semantic priors in a single efficient pipeline. Built on an RF-DETR/LW-DETR-style foundation with a DINOv2 backbone, HOIBlender removes the dedicated transformer encoder and selects HOI candidates directly from multi-scale projector outputs. Its dual-stage decoder first stabilizes human-object geometry and then performs verb and HOI prediction with progressive BLIP-2 prior fusion at both the decoder-memory and classifier-weight levels, while grouped-query training enriches optimization without increasing inference cost. On HICO-DET, HOIBlender consistently outperforms recent SOV-style baselines across three model scales while preserving competitive latency and parameter budgets. These results suggest that detection efficiency and deep vision-language integration need not be traded against each other, offering a practical recipe for future HOI detectors targeting both accuracy and deployment.

{\small
\bibliographystyle{plainnat}
\bibliography{references}
}

\end{document}